\documentclass[10pt,twocolumn,letterpaper]{article}

\usepackage{iccv}              

\definecolor{iccvblue}{rgb}{0.21,0.49,0.74}
\usepackage[pagebackref,breaklinks,colorlinks,allcolors=iccvblue]{hyperref}

\usepackage{newfloat}
\usepackage{listings}
\usepackage{adjustbox}
\usepackage{multirow}
\usepackage{booktabs}
\usepackage{color}
\usepackage{graphicx}
\usepackage{subcaption}
\usepackage{caption}
\usepackage{amsmath}
\def\paperID{2163} 
\def\confName{ICCV}
\def\confYear{2025}

\title{Pseudo-Label Quality Decoupling and Correction for \\ Semi-Supervised Instance Segmentation}

\author{Jianghang Lin\textsuperscript{\rm 1}
,
Yilin Lu\textsuperscript{\rm 1}
,
Yunhang Shen\textsuperscript{\rm 2}
,
Chaoyang Zhu\textsuperscript{\rm 1}
,
Shengchuan Zhang\textsuperscript{\rm 1}
,
Liujuan Cao\textsuperscript{\rm 1}\thanks{Corresponding Author}
,
Rongrong Ji\textsuperscript{\rm 1}
\\
\textsuperscript{\rm 1}Key Laboratory of Multimedia Trusted Perception and Efficient Computing, Ministry of Education of China,\\Xiamen University, China.
\\
\textsuperscript{\rm 2}Tencent Youtu Lab, China.
\\
{\tt\small \{hunterjlin007,yilinlu\}@stu.xmu.edu.cn}, {\tt\small \{zsc\_2016,caoliujuan,rrji\}@xmu.edu.cn},\\
{\tt\small \{sean.zhuh,shenyunhang01\}@gmail.com}
}

\begin{document}
\maketitle
\begin{abstract}
Semi-Supervised Instance Segmentation~(SSIS) involves classifying and grouping image pixels into distinct object instances using limited labeled data.
This learning paradigm usually faces a significant challenge of unstable performance caused by noisy pseudo-labels of instance categories and pixel masks.
We find that the prevalent practice of filtering instance pseudo-labels assessing both class and mask quality with a single score threshold, frequently leads to compromises in the trade-off between the qualities of class and mask labels.
In this paper, we introduce a novel \textbf{P}seudo-\textbf{L}abel Quality \textbf{D}ecoupling and \textbf{C}orrection~(\textbf{PL-DC}) framework for SSIS to tackle the above challenges.
Firstly, at the instance level, a decoupled dual-threshold filtering mechanism is designed to decouple class and mask quality estimations for instance-level pseudo-labels, thereby independently controlling pixel classifying and grouping qualities.
Secondly, at the category level, we introduce a dynamic instance category correction module to dynamically correct the pseudo-labels of instance categories, effectively alleviating category confusion.
Lastly, we introduce a pixel-level mask uncertainty-aware mechanism at the pixel level to re-weight the mask loss for different pixels, thereby reducing the impact of noise introduced by pixel-level mask pseudo-labels.
Extensive experiments on the COCO and Cityscapes datasets demonstrate that the proposed PL-DC achieves significant performance improvements, setting new state-of-the-art results for SSIS.
Notably, our PL-DC shows substantial gains even with minimal labeled data, achieving an improvement of $+11.6$ $m$AP with just $1\%$ COCO labeled data and $+15.5$ $m$AP with $5\%$ Cityscapes labeled data.
%
%
\end{abstract}    
\section{Introduction}
\label{sec:intro}
\begin{figure}[!t]
\centering
\includegraphics[width=0.85\linewidth]{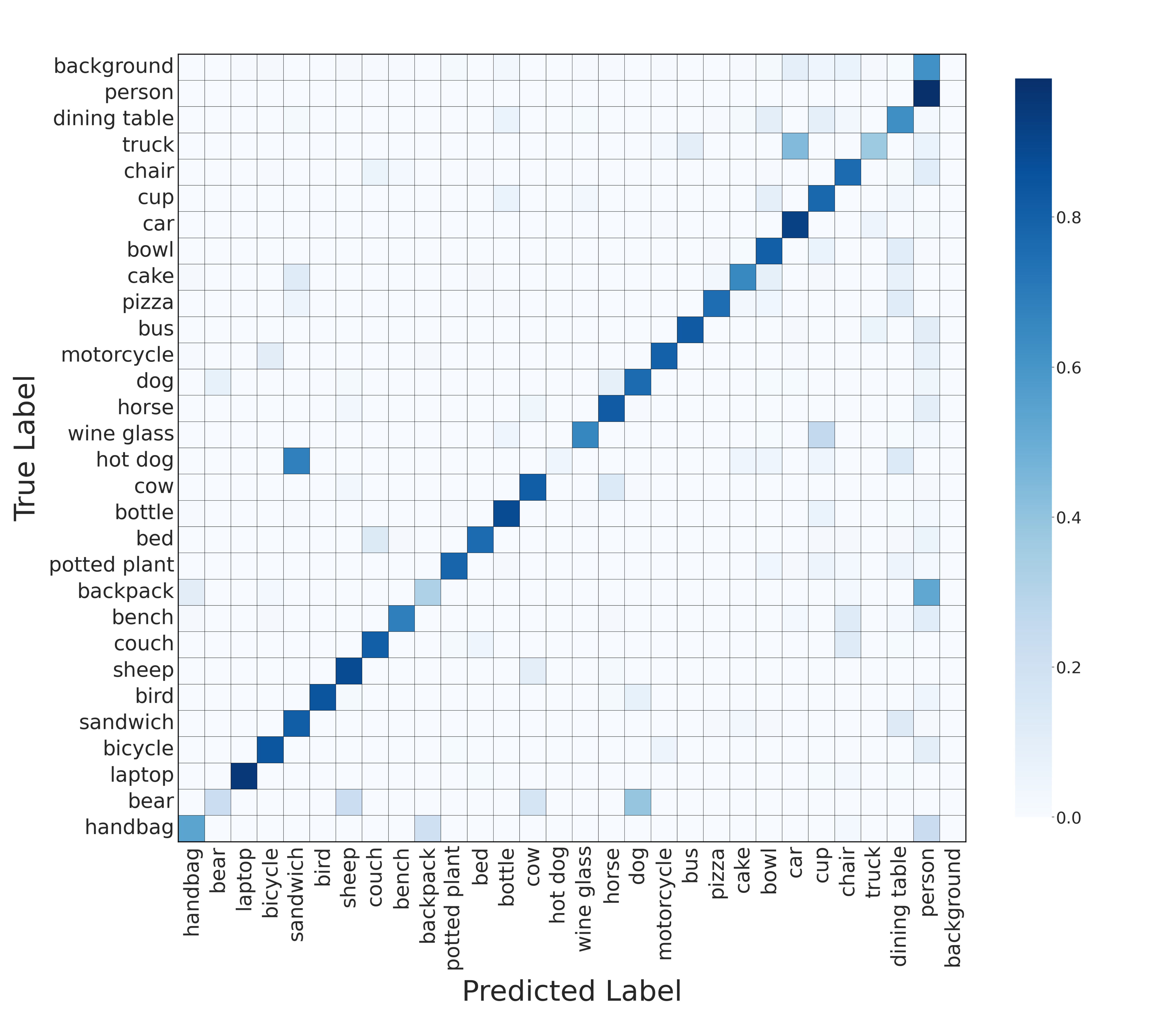}
\vspace{-15pt}
\caption{
\textbf{Confusion matrix of the model trained on $1\%$ COCO.}
For clarity, we visualize only the $29$ most confused object categories and $1$ background category. 
}
\label{fig:confusion_matrix}
\vspace{-20pt}
\end{figure}

Artificial intelligence community has witnessed significant progress in object instance segmentation in the past decades, especially with the popularity of deep learning.
\begin{figure}[!t]
\centering
\includegraphics[width=1.0\linewidth]{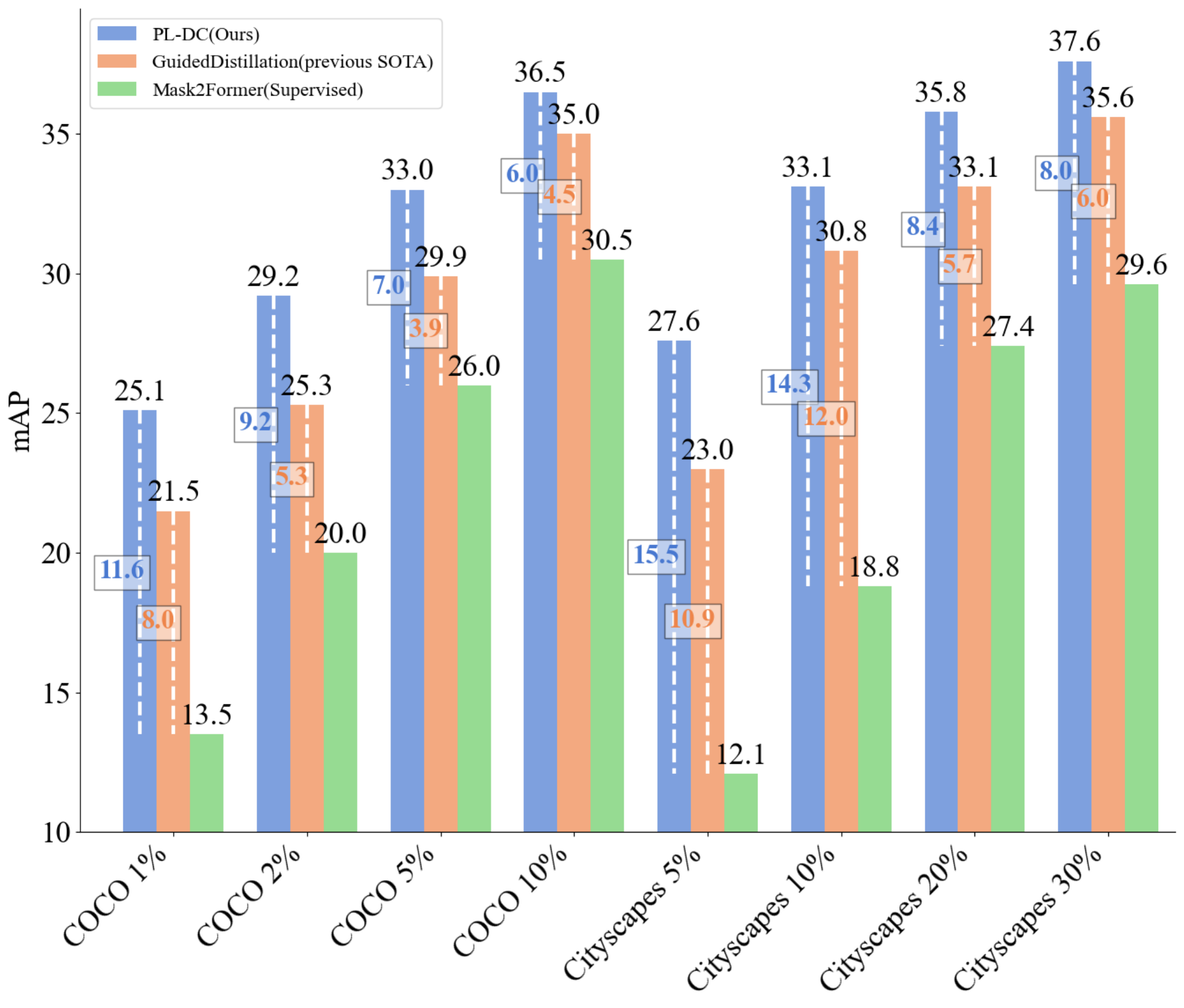}
\vspace{-25pt}
\caption{
    The proposed PL-DC outperforms the previous state-of-the-art SSIS method, GuidedDistillation~\cite{berrada2023guided}, across all settings.
    Moreover, PL-DC achieves significant improvements compared to the fully-supervised Mask2Former.
}
\label{fig:histogram}
\vspace{-22pt}
\end{figure}
%
Large-scale human-annotated datasets such as COCO~\cite{lin2014microsoft}, LVIS~\cite{Gupta2019}, Cityscapes~\cite{cordts2016cityscapes} and BDD100K~\cite{yu2020bdd100k} have been published to study fully-supervised instance segmentation~(FSIS) at the pixel level, leading significant improvement in image understanding.
Nevertheless, the laborious and lavish collection of pixel-level annotations has severely barricaded the applicability of FSIS in practical application.
Semi-supervised learning has emerged to exploit large-scale unlabeled data in image classification and object detection to improve performance, given limited labeled data.
%
\begin{figure*}[ht!]
    \centering
    \includegraphics[width=1.0\linewidth]{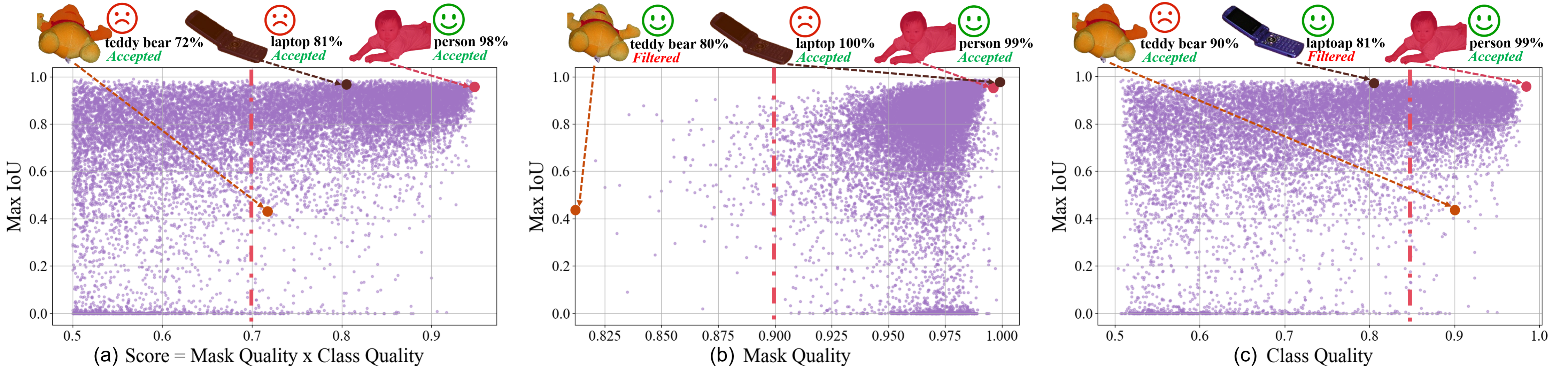}
    \vspace{-25pt}
    \caption{
    The relationship between predicted instance scores and the IoU of predicted versus ground-truth instance masks.
    (a) Predicted instance scores, derived from the product of mask quality and class quality, do not show a positive correlation with IoU. (b) Decoupled mask quality and (c) class quality independently influence the segmentation and classification outcomes of pseudo-labels.
    }
    \label{fig:pseudo_label_threshold}
    \vspace{-20pt}
\end{figure*}
%
However, the instance segmentation task is more challenging than classification and object detection tasks, which not only learns semantic-level categories and instance-level coordinates but also requires pixel-level classification and grouping.
Therefore, semi-supervised instance segmentation~(SSIS) still lies far behind semi-supervised image classification and object detection.
We conclude three main challenges that hinder the development of SSIS: 
(1) At the instance level, filtering pseudo-labels with a coupled score threshold fails to evaluate the class and mask qualities of instances simultaneously.
As shown in Fig.\,\ref{fig:pseudo_label_threshold} (a), we find that the predicted instance scores are not positively correlated to the IoU with the ground-truth instances, which may lead to bias estimation of pseudo-labels.
(2) At the category level, categories with similar appearance or frequently co-occurring are prone to category prediction confusion, as shown in Fig.\,\ref{fig:confusion_matrix}.
For example, cars are mistaken for trucks due to their similar structure. Bears are mistaken for Dogs because they look similar to dogs.
Hot dogs often appear with sandwiches simultaneously, which confuses the models.
(3) At the pixel level, pseudo-labels of dense masks are usually imperfect compared to those of one-hot categories.
This is because the pixel-level mask loss calculates all pixels of the entire image, while the instance-level classification loss only calculates a relatively small number of instances.
Obviously, the number of pixel-level mask pseudo-labels is much larger than that of instance-level category pseudo-labels, which makes model training more vulnerable to pixel-level mask pseudo-labels.
Towards solving the above three problems, we present a new semi-supervised instance segmentation framework, referred to as pseudo-label quality decoupling and correction~(PL-DC).
We innovate in three aspects:
(1) Observations from Fig.\,\ref{fig:pseudo_label_threshold} (b)(c) suggest that decoupled class and mask estimation independently control the quality of classifying and grouping.
To capitalize on this, we propose a decoupled dual-threshold filtering mechanism.
This approach ensures instance-level pseudo-labels have both high qualities of class and mask, thus eliminating the detrimental effects of potential trade-offs between class quality and mask quality inherent in traditional coupled score threshold filtering mechanisms.
(2) We introduce a dynamic instance category correction module leveraging the visual-language alignment model, CLIP, which has been pre-trained on large-scale image-text pairs.
This module dynamically corrects the probability distribution of category pseudo-labels, effectively mitigating category confusion.
Specifically, CLIP processes image patches extracted from the filtered mask pseudo-labels and the text descriptions of all categories to compute a similarity probability distribution.
Combined with the predicted category pseudo-label’s probabilities by teacher model, the adjusted distribution updates category assignments based on the highest probability.
(3) We implement a pixel-level mask uncertainty-aware loss function, which assigns small loss weights to regions with high uncertainty in mask pseudo-labels and large weights to areas with low uncertainty.
This loss function reduces the influence of noise prevalent in pixel-level mask pseudo-labels, enhancing model robustness and accuracy.

Extensive experiments on COCO and Cityscapes demonstrate that our PL-DC achieves new state-of-the-art results.
Specifically, on COCO with 1\%, 2\%, 5\%, 10\%, and 100\% labeled images, our proposed PL-DC achieves significant performance boosts with increases of +11.6\%, +9.2\%, +7.0\%, +6.0\%, and +5.3\% $m$AP, respectively.
On the Cityscapes dataset, with 5\%, 10\%, 20\%, and 30\% labeled images, PL-DC also exhibits substantial enhancements, recording $m$AP improvements of +15.5\%, +14.3\%, +8.4\%, and +8.0\%, respectively.
\section{Related Work}
\label{sec:related_work}
\subsection{Instance Segmentation}
\label{sec:rw_is}
Instance segmentation is a crucial task in computer vision that classifies each pixel in an image into distinct object instances, identifying not only the object category but also differentiating between multiple objects of the same category. 
This detailed understanding is essential for applications like autonomous driving and medical imaging.
Current approaches can be grouped into three categories: detection-based, clustering-based, and query-based methods.
\textit{Detection-based} methods~\cite{he2017mask,bolya2019yolact,cai2019cascade,li2022exploring} extend traditional object detection frameworks by first generating bounding boxes and then classifying pixels within those boxes to segment objects.
A prominent example is Mask R-CNN~\cite{he2017mask}, which builds on the Faster R-CNN~\cite{girshick2015fast} framework by adding a segmentation branch to predict masks for each Region of Interest (RoI). This approach has been highly influential and remains a benchmark for many later works.
\textit{Clustering-based} methods~\cite{de2017semantic,gao2019ssap,liu2017sgn,bai2017deep} group pixels based on their features and spatial proximity to form object instances.
Techniques like Mean Shift~\cite{comaniciu2002mean} or Graph Cut~\cite{boykov2001interactive} are commonly used for pixel clustering. A representative method is the Deep Watershed Transform~\cite{bai2017deep}, which interprets an image as a topographic surface and applies watershed algorithms to segment instances using learned energy functions.
\textit{Query-based} methods leverage learnable query embeddings to directly segment instances, often utilizing transformer architectures for enhanced accuracy and efficiency.
DETR~\cite{carion2020end}, for example, employs a transformer encoder-decoder architecture with bipartite matching and a set-based loss function to predict class labels and masks for each object instance in an end-to-end manner. 
MaskFormer~\cite{cheng2021per} and Mask2Former~\cite{cheng2021mask2former} further improve DETR by integrating semantic, instance, and panoptic segmentation into a unified framework. MaskFormer uses a transformer decoder to predict segmentation masks directly, while Mask2Former enhances it with a multi-scale design, masked attention, and improved mask prediction.
\subsection{Semi-Supervised Instance Segmentation}
\label{sec:rw_ssis}
Semi-supervised learning aims to reduce the dependency on labeled data by incorporating unlabeled data during training.
It has made significant progress in image classification and object detection tasks with techniques such as self-training, consistency regularization, and adversarial learning.
Pseudo-label-based methods~\cite{bachman2014learning,liu2021unbiased,sohn2020simple,xu2021end,mi2022active} leverage pre-trained models to generate annotations for unlabeled images, which are then used to train the model. 
Consistency-regularization-based methods~\cite{berthelot2019semi,berthelot2019mixmatch,gao2020consistency,jeong2019consistency} incorporate various data augmentation techniques, such as random regularization and adversarial perturbation, to generate different inputs for a single image and enforce consistency between these inputs during training.
In the instance segmentation task, addressing pixel-level noise presents greater challenges compared to image-level classification and box-level detection, leading to slower advancements in this area.
Noisy Boundary~\cite{wang2022noisy} was the first to formally introduce the semi-supervised instance segmentation task.
It assumes that noise exists in the boundary area of the object and effectively utilizes the noise boundary information in unlabeled images and pseudo-labels to improve instance segmentation performance by combining a noise-tolerant mask terminator and a boundary-preserving map.
Instead of static pseudo-label generation, Polite Teacher~\cite{filipiak2024polite} uses dynamic pseudo-label generation built on the Teacher-Student mutual learning framework with a single-stage anchor-free detector, CenterMask~\cite{lee2020centermask}, and utilizes confidence thresholding for bounding boxes and mask scoring to filter out noisy pseudo-labels. 
In contrast to filtering out pseudo-labels with low confidence, PAIS~\cite{Hu2023pseudolabel} leverages them by using a dynamic aligning loss that adjusts the weights of semi-supervised loss terms based on varying class and mask score pairs.
Unlike our Pixel-Level Mask Uncertainty-Aware, PAIS introduces an IoU prediction branch that alters the original architecture of the instance segmentation model.
Different from the previous focus on detection-based Mask R-CNN~\cite{he2017mask}, GuidedDistillation~\cite{berrada2023guided} proposed a three-stage semi-supervised Teacher-Student distillation framework and used a powerful query-based instance segmentation model Mask2Former~\cite{cheng2021mask2former} for the first time, achieving promising performance improvements.
Despite its effectiveness, it remains constrained by the limitations of coupled score filtering of pseudo-labels.
\begin{figure*}[!t]
\centering
\includegraphics[width=1.0\textwidth]{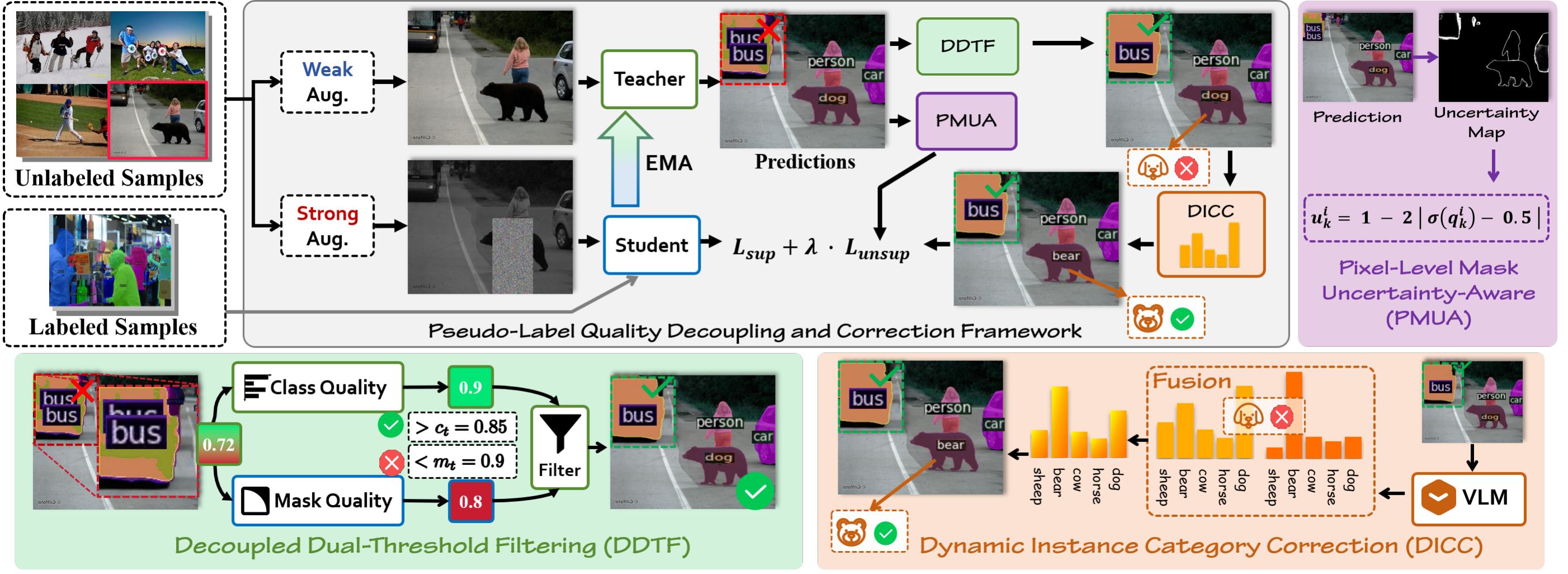}
\vspace{-20pt}
\caption{\textbf{Framework of our proposed pseudo-label quality decoupling and correction~(PL-DC) for semi-supervised instance segmentation.}
PL-DC includes two segmentation models, both Mask2Former~\cite{cheng2021mask2former}, with identical configurations, namely Teacher and Student.
The Teacher model generates an uncertainty map for Pixel-Level Mask Uncertainty-Aware training, filters pseudo-labels by the Decoupled Dual-Threshold Filtering~(DDTF) mechanism, and further corrects category by Dynamic Instance Category Correction~(DICC).
The Teacher's parameters are gradually updated from the Student model via Exponential Moving Average~(EMA).
The Student is trained using both ground-truth labels and pseudo-labels~(with uncertainty map), denoted as $\mathcal{L}_\mathrm{sup}$ and $\mathcal{L}_\mathrm{unsup}$, respectively.
}
\label{fig:framework}
\vspace{-15pt}
\end{figure*}
\section{Method}
\label{sec:method}
In this section, we outline our PL-DC framework designed to tackle the three challenges commonly encountered in semi-supervised instance segmentation, as depicted in Fig.\,\ref{fig:framework}.
The objective is to leverage both labeled data $D_L=\{X_L, Y_L\}$ and unlabeled data $D_U=\{X_U\}$ to optimize instance segmentation performance, where $X$ represents image samples and $Y$ denotes mask annotations with their corresponding classes.
Our framework utilizes a teacher-student structure in semi-supervised learning.
It incorporates two instance segmentation networks with identical structures: one acting as the teacher and the other as the student.
The teacher network generates pseudo-labels for the unlabeled data, which the student network uses to learn alongside the labeled data.
Consequently, the overarching loss function is defined as:
\begin{equation}
    \mathcal{L} = \mathcal{L}_\mathrm{sup} + \lambda \mathcal{L}_\mathrm{unsup},
    \label{eq:total_loss}
\end{equation}
where $\mathcal{L}_\mathrm{sup}$ and $\mathcal{L}_\mathrm{unsup}$ denote the losses for supervised and unsupervised learning, respectively, and $\lambda$ is a hyper-parameter that balances these losses.
For instance segmentation, the supervised learning loss is defined as:
\begin{equation}
    \mathcal{L}_\mathrm{sup} = \mathcal{L}_\mathrm{cls} + \mathcal{L}_\mathrm{mask},
    \label{eq:sup_loss}
\end{equation}
with $\mathcal{L}_\mathrm{cls}$ is the classification cross-entropy loss and $\mathcal{L}_\mathrm{mask}$ is the pixel-level binary cross-entropy loss, which may optionally include dice loss.
The unsupervised learning loss mirrors Eq.\,\ref{eq:sup_loss}, but the supervision comes from the pseudo-labels generated by the teacher network.
The student network updates its parameters via stochastic gradient descent~(SGD). 
To prevent overfitting, the teacher network's gradients are frozen, and its parameters are updated from the student network using the Exponential Moving Average~(EMA)~\cite{tarvainen2017mean}.
\subsection{Decoupled Dual-Threshold Filtering}
\label{sec:method_ddtf}
We take Mask2Former as our foundational instance segmentation network structure due to its powerful performance in the instance segmentation field. 
This model computes the instance score $\mathbf{s}_k$ as the product of the class quality $\mathbf{c}_k$ and mask quality $\mathbf{m}_k$. 
Class quality $\mathbf{c}_k$ is defined as:
\begin{equation}
    \mathbf{c}_k = \frac{e^{x_k^j}}{\sum_{i=1}^{N} e^{x_k^i}},
\end{equation}
where $N$ is the number of class, $x_k^i$ represents the logit of class $i$ prediction for the $k$-th instance, and $x_k^j$ is the maximum class logit.
Mask quality $\mathbf{m}_k$ is calculated using the following formula:
\begin{equation}
    \mathbf{m}_k = \frac{\sum_{i=1}^{HW}{\sigma(q_k^i) \times \textbf{1}[\sigma(q_k^i)>0.5]}}{\sum_{i=1}^{HW}{\textbf{1}[\sigma(q_k^i)>0.5]}},
\end{equation}
where $HW$ is the total number of pixels in the mask, $q_k^i$ is the per-pixel logit of the predicted mask for the $k$-th instance, $\sigma$ denotes the sigmoid function, and $\textbf{1}[\sigma(q_k^i)>0.5]$ is an indicator function that equals $1$ if $\sigma(q_k^i)>0.5$, and $0$ otherwise.
In fully-supervised learning, the availability of ample labeled data allows the model to effectively and comprehensively increase both $\mathbf{c}_k$ and $\mathbf{m}_k$, thereby $\mathbf{s}_k=\mathbf{c}_k \cdot \mathbf{m}_k$ accurately measuring the quality of each instance.
However, in semi-supervised learning, the limited labeled data and the presence of noisy pseudo-labels from unlabeled data mean that $\mathbf{c}_k$ and $\mathbf{m}_k$ cannot always be optimized well simultaneously.
This limitation results in the coupled instance score $\mathbf{s}_k$ sometimes failing to reflect the true quality of the instance accurately.
Using a single threshold for coupled instance score $\mathbf{s}_k$ to filter pseudo-labels~\cite{berrada2023guided} leads to a competitive relationship between class quality $\mathbf{c}_k$ and mask quality $\mathbf{m}_k$. 
For instance, an instance score $\mathbf{s}_k$ of $0.72$ could result from either a $\mathbf{m}_k$ of $0.8$ and a $\mathbf{c}_k$ of $0.9$, or a $\mathbf{m}_k$ of $0.96$ and a $\mathbf{c}_k$ of $0.75$.
If we use an instance score threshold of $0.7$, the model may not adequately account for both $\mathbf{c}_k$ and $\mathbf{m}_k$, leading to pseudo-labels that are sometimes misclassified or possess poor mask quality.
We observe that the decoupled class quality $\mathbf{c}_k$ and mask quality $\mathbf{m}_k$ independently control the quality of instance pseudo-labels, as illustrated in Fig.\,\ref{fig:pseudo_label_threshold} (b)(c). 
Based on this insight, we propose a Decoupled Dual-Threshold Filtering~(DDTF) mechanism, which effectively mitigates the competition between $\mathbf{c}_k$ and $\mathbf{m}_k$.
Specifically, the teacher network processes the weakly augmented image $X_U^{weak}$ as input and generates $Q$ predicted instance results ${(C_i,M_i)}_{i=1,2,...,Q}$.
These predictions are then selectively filtered based on both the mask quality threshold $m_t$ and the class quality threshold $c_t$, as formulated below:
\begin{equation}
    \hat{Y}_U = \{(C_i, M_i) \mid C_i \geq c_t \& M_i \geq m_t, \, i = 1,...,Q\}.
\end{equation}
It is worth noting that the dual-threshold filtering method has been used in PAIS~\cite{Hu2023pseudolabel}. 
However, we emphasize that the mask quality assessment in PAIS is achieved by modifying the original model structural, adding a mask IoU prediction branch to learn the mask quality from scarce labeled data.
This approach has been proven effective in fully-supervised MS-RCNN~\cite{huang2019mask}.
In semi-supervised instance segmentation, however, the scarcity of labeled data leads to inaccurate mask IoU predictions, which cannot effectively measure the quality of predicted masks on unlabeled data. 
This is demonstrated in ‘a’ and ‘b’ of Tab.\,\ref{table_general_capabilities}, where mask IoU prediction overfits with only $1\%$ labeled data, failing to evaluate mask quality. 
%
%
In contrast, our DDTF does not require changes to the model structure. It assesses mask quality based on foreground pixel uncertainty in the predicted masks, thus avoiding the impact of overfitting.
\subsection{Dynamic Instance Category Correction}
\label{sec:method_dicc}
Ideally, semi-supervised learning addresses the challenge of label scarcity.
However, it is often compromised by inherent imbalances in instance segmentation.
For instance, in the COCO dataset, \textit{person} instances make up $30\%$ of all foreground training instances, while \textit{hair driers} and \textit{toasters} represent only $0.023\%$ and $0.026\%$, respectively.
Such disparities lead the model to favor predicting dominant classes, especially when training data is limited, resulting in a bias towards these categories.
This exacerbates the imbalance in the generated pseudo-labels, leading to severe prediction biases during training.
As depicted in Fig.\,\ref{fig:confusion_matrix}, instances that look similar or frequently co-occur are prone to category prediction confusion by a close dominant category.
For example, bears are often mistaken for dogs due to their similar appearance, and hot dogs are frequently confused with sandwiches due to common co-occurrences.
In recent years, large visual-language alignment models~(LVLMs) pre-trained on extensive image-text pairs have shown exceptional zero-shot classification capabilities. 
Many works have already leveraged LVLMs to explore open-vocabulary~\cite{yao2023detclipv2,zhou2022detecting}, weakly supervised~\cite{lin2024weakly} and semi-supervised object detection~\cite{haghighi2023pseudo}. However, to the best of our knowledge, no one has yet explored the potential of LVLMs in semi-supervised instance segmentation.
We believe these models can effectively address the inaccuracies in pseudo-labels.
To leverage this advantage, we propose Dynamic Instance Category Correction~(DICC) to rectify the categories of pseudo-labels after DDTF filtering.
For simplicity, we utilize CLIP~\cite{radford2021learning} as a representative visual-language alignment model for our DICC. 
Specifically, for each pseudo-label $(C_i, M_i)$, CLIP processes the image patch $x_i^{pool}$, extracted from $M_i$, alongside the textual descriptions of all categories $\mathbf{t} \in R^N$ from the training set.
A probability distribution $p_i^{clip} \in R^N$ is computed as follows:
\begin{equation}
    p_i^{clip} = Softmax(CLIP_V(x_i^{pool}) \cdot CLIP_T(\mathbf{t})),
\end{equation}
where $CLIP_V$ and $CLIP_T$ are the vision and text encoders of CLIP, respectively.
We then dynamically fuse the probability distribution $\hat p_i$ from the teacher model’s predictions for $C_i$ with $p_i^{clip}$ to create a final distribution $p_i^{f} \in R^N$.
The category with the highest score in $p_i^{f}$ is selected as the corrected category pseudo-label $C_i^{corr}$:
\begin{equation}
    w = 0.25(\cos({\frac{it\_cur}{it\_max}\pi})+1),
    \label{eq:dicc_1}
\end{equation}
\begin{equation}
    p_i^{f} = w \cdot p_i^{clip} + (1-w) \cdot \hat p_i,
    \label{eq:dicc_2}
\end{equation}
\begin{equation}
    C_i^{corr} = \arg\max(p_i^{f}),
    \label{eq:dicc_3}
\end{equation}
where $it\_cur$ and $it\_max$ represent the current and maximum training iterations, respectively. 
The weighting factor $w$ decays from $0.5$ to $0$ as the teacher model's accuracy improves, reflecting its increasing reliability.
%
This dynamic approach effectively balances the strengths of both the teacher model and the LVLM, allowing them to complement each other’s ability to recognize unfamiliar categories.
For more analysis, see Appendix~\ref{sec_sup:impact_clip}.
\subsection{Pixel-Level Mask Uncertainty-Aware}
\label{sec:method_plmua}
In instance segmentation, the loss function for model training typically includes an instance-level classification cross-entropy loss and a pixel-level mask binary cross-entropy loss.
The pixel-level mask loss considers all pixels in the entire image, whereas the instance-level classification loss is concerned with a relatively smaller number of instances.
Consequently, the number of pixel-level mask pseudo-labels significantly exceeds that of instance-level category pseudo-labels, making the model training more susceptible to the influence of pixel-level mask pseudo-labels.
Given the extensive use of pixel-level mask pseudo-labels in semi-supervised learning, it is crucial to account for the uncertainty associated with these labels.
Recent work Noisy Boundaries~\cite{wang2022noisy} introduced the Boundary-preserving Map~(BMP), which re-weights the mask loss for different pixels based on their proximity to object boundaries, thereby making model training sensitive to uncertain mask pixels.
Noisy Boundaries posits that uncertainty primarily exists at object boundaries. 
However, we have observed significant uncertainty in areas where multiple objects overlap, a scenario where BMP is less effective.
To address this broader range of uncertainties, we propose the Pixel-level Mask Uncertainty-Aware~(PMUA) approach to re-weight the mask loss across different pixels comprehensively.
We define the uncertainty $u_k^i$ of the per-pixel mask as:
\begin{equation}
    u_k^i = 1 - 2 \left| \sigma (q_k^i) - 0.5 \right|,
\end{equation}
where $\sigma (q_k^i)$ is the predicted foreground per-pixel binary mask probability of the $k$-th instance by the teacher model.
Following the DDTF and DICC processes, we obtain corrected pseudo-labels $\hat{Y}_U^{corr} = \{(C_k^{corr}, M_k, u_k) \mid C_k^{corr} \in \{1, \dots ,N\}, M_k \in \{0,1\}^{HW}, u_k \in [0,1]^{HW} \}_{k=0}^{N^{pgt}}$ for unlabeled data, where $C_k^{corr}$ is the corrected pseudo ground truth class labels, $M_k$ is the pseudo ground truth binary mask, and $u_k$  represents the uncertainty values for each $M_k$, $N^{pgt}$ is the total number of pseudo-labeled instances obtained.
Then, pixel-level mask binary cross-entropy loss for unlabeled data to train the student model is defined as:
\begin{equation}
    \begin{split}
        \mathcal{L}_{mask}^{unsup} =& -\frac{1}{QHW} \sum_{k=1}^{Q} \sum_{i=1}^{HW} ( 1 - u_{\hat{\sigma}(k)}^i ) [ M_{\hat{\sigma}(k)}^i \log (t_k^i) \\ +
        & (1 - M_{\hat{\sigma}(k)}^i) \log (1 - t_k^i) ],
    \end{split}
\end{equation}
where $\hat{\sigma}$ is the optimal assignment calculated using the Hungarian algorithm, $t_k^i$ is the predicted foreground per-pixel binary mask probability of the $k$-th instance by student model.
In Appendix~\ref{sec_sup:PMUA_Proof}, we derive the gradient of $\mathcal{L}_{mask}^{unsup}$ with respect to the student model parameters $\theta$ and prove that a higher $u_{\hat{\sigma}(k)}^i$, indicating greater noise in $M_{\hat{\sigma}(k)}^i$, results in a proportionally lesser influence of the pixel's pseudo-label on the update of $\theta$,  thereby improving the robustness of the training process under label uncertainty.
\section{Experiments}
\label{sec:exp}
\subsection{Settings and Implementation Details}
\textbf{Experimental Settings.}
We benchmark our proposed PL-DC on COCO~\cite{lin2014microsoft} and Cityscapes~\cite{cordts2016cityscapes} datasets following existing works~\cite{berrada2023guided,Hu2023pseudolabel,wang2022noisy}.
The COCO dataset, which comprises $80$ categories, is notably challenging for instance segmentation.
It includes $118k$ \textit{train2017} labeled images, $5k$ \textit{val2017} labeled images and $123k$ \textit{unlabel2017} unlabeled images.
We randomly sample $1\%$, $2\%$, $5\%$, and $10\%$ of the images from the \textit{train2017} split as labeled data and treated the rest as unlabeled data following common settings.
Additionally, we utilized the entire \textit{train2017}, denoted as $100\%$, as labeled data and incorporated the \textit{unlabel2017} as unlabeled data for PL-DC evaluation.
The Cityscapes dataset contains $2,975$ training images and $500$ validation images of size $1024 \times 2048$ taken from a car driving in German cities, labeled with $8$ semantic instance categories.
We follow \cite{berrada2023guided} sample $5\%$, $10\%$, $20\%$, and $30\%$ of the images from the training set as labeled images and treat the rest as unlabeled ones.
We conducted evaluations using the COCO \textit{val2017} and the Cityscapes validation sets for their respective experimental settings, reporting the standard COCO $m$AP metric as in previous studies.

\noindent\textbf{Implementation Details.}
We employ Mask2Former~\cite{cheng2021mask2former} with ResNet-50 as our baseline instance segmentation network, and the implementation and hyper-parameters setting are the same as those in Detectron2~\cite{wu2019detectron2}.
By default, all experiments are conducted on a single machine equipped with four $3090$ GPUs, each with $24$ GB of memory.
For optimization, we utilize AdamW~\cite{loshchilov2017decoupled} with a learning rate and weight decay both set at $0.0001$. Due to limited GPU memory, all network backbones are frozen.
Following~\cite{liu2021unbiased}, we apply random horizontal flip and scale jittering as weak augmentations for the teacher model, while the student model receives strong augmentations including horizontal flip, scale jittering, color jittering, grayscale, gaussian blur, and CutOut~\cite{devries2017improved}.
We use mask quality threshold $m_t=0.9$ and class quality threshold $c_t=0.85$ to filter the pseudo-labels.
We use $\alpha=0.9996$ for EMA and $\lambda=1$ for the unsupervised loss $\mathcal{L}_\mathrm{unsup}$.
For the COCO setup, we pre-train the teacher model with the supervised learning defined in Eq.\,\ref{eq:sup_loss} about $20$k iterations.
Afterward, the student model is initialized with the parameters of the teacher model. 
The total training iterations for each semi-supervised learning are all $360K$~($50$ epochs), with batch sizes consistently comprising $8$ labeled and $8$ unlabeled images unless otherwise specified.
For Cityscapes setup, the hyper-parameters mirror those of the COCO configuration, except the total training duration is reduced to $180k$ iterations, and the batch sizes are halved to $8$.
For more implementation details, see Appendix~\ref{sec_sup:implementation_details}.
\subsection{Comparison with Other Methods}
In Tab.\,\ref{table_sota_coco}, We compare our PL-DC with other semi-supervised instance segmentation frameworks on the COCO dataset.
Our observations reveal that PL-DC consistently outperforms the current state-of-the-art method, GuidedDistillation~\cite{berrada2023guided}, across all COCO-labeled data ratios.
Notably, our PL-DC shows a more substantial increase in $m$AP at lower labeled data ratios compared to the fully supervised Mask2Former.
Specifically, the $m$AP improvements are $+11.6$, $+9.2$, $+7.0$, and $+6.0$ for $1\%$, $2\%$, $5\%$, and $10\%$ labeled data, respectively, underscoring PL-DC's effective use of large-scale unlabeled data.
In contrast, GuidedDistillation exhibits smaller and somewhat counterintuitive $m$AP gains of $+3.9$ at $5\%$ and $+4.5$ at $10\%$, indicating a higher dependency on labeled data.
Moreover, employing $100\%$ of the COCO labeled data, PL-DC further achieves an enhancement of $+5.3$ $m$AP by integrating $123k$ \textit{unlabel2017} COCO images.
To evaluate the generalizability of our PL-DC, we conducted experiments on the Cityscapes autonomous driving dataset, which features a larger resolution closer to industrial practicality.
As shown in Tab.\,\ref{table_sota_cityscapes}, PL-DC continues to outperform under varied labeled data proportions. 
Specifically, compared with Supervised Mask2Former, our PL-DC improved $m$AP by $+15.5$, $+14.3$, $+8.4$, and $+8.0$ at $5\%$, $10\%$, $20\%$, and $30\%$ labeled data, respectively, while GuidedDistillation still exhibited counterintuitive results at $5\%$ and $10\%$ labeled data. 
These results confirm that our PL-DC is robust and can be effectively generalized across different datasets.
\begin{table}[t]
    \begin{center}
        \begin{adjustbox}{max width=1.0\linewidth}
            \begin{tabular}{l|lllll}
                \toprule
                
                Method & 1\% & 2\% & 5\% & 10\% & 100\%\\
                
                \midrule

                Mask-RCNN, Superised & 3.5 & 9.3 & 17.3 & 22.0 & 34.5\\
                
                Mask2Former, Superised & 13.5 & 20.0 & 26.0 & 30.5 & 43.5\\
                
                \midrule

                DD~\cite{Ilija2018Data} & 3.8 & 11.8 & 20.4 & 24.2 & 35.7\\ 
                Noisy Boundaries~\cite{wang2022noisy} & 7.7 & 16.3 & 24.9 & 29.2 & 38.6\\ 
                Polite Teacher~\cite{filipiak2024polite} & 18.3 & 22.3 & 26.5 & 30.8 & -\\ 
                PAIS~\cite{Hu2023pseudolabel} & 21.1 & - & 29.3 & 31.0 & 39.5\\ 
                GuidedDistillation~\cite{berrada2023guided} & 21.5~\small{\color{blue}({+8.0})} & 25.3~\small{\color{blue}({+5.3})} & 29.9~\small{\color{blue}({+3.9})} & 35.0~\small{\color{blue}({+4.5})} & -\\
                
                \midrule

                PL-DC~(Ours) & \textbf{25.1}~\small{\color{blue}({+11.6})} & \textbf{29.2}~\small{\color{blue}({+9.2})} & \textbf{33.0}~\small{\color{blue}({+7.0})} & \textbf{36.5}~\small{\color{blue}({+6.0})} & \textbf{48.8}~\small{\color{blue}({+5.3})} \\
                \bottomrule
            \end{tabular}
        \end{adjustbox}
    \end{center}
    \vspace{-5pt}
    \caption{
         Comparison with other SSIS on COCO.
    }
    \label{table_sota_coco}
    \vspace{-10pt}
\end{table}

\begin{table}[t]
    \begin{center}
        \begin{adjustbox}{max width=1.0\linewidth}
            \begin{tabular}{l|llll}
                \toprule
                
                Method & 5\% & 10\% & 20\% & 30\%\\
                
                \midrule
                
                Mask-RCNN, Supervised & 11.3 & 16.4 & 22.6 & 26.6\\

                Mask2Former, Supervised & 12.1 & 18.8 & 27.4 & 29.6\\
                \midrule
                
                DD~\cite{Ilija2018Data} & 13.7 & 19.2 & 24.6 & 27.4\\
                
                STAC~\cite{sohn2020simple} & 11.9 & 18.2 & 22.9 & 29.0\\
                
                CSD~\cite{JisooJeong2019ConsistencybasedSL} & 14.1 & 17.9 & 24.6 & 27.5\\
                
                CCT~\cite{YassineOuali2020SemiSupervisedSS} & 15.2 & 18.6 & 24.7 & 26.5\\
                
                Dual-branch~\cite{WenfengLuo2020SemisupervisedSS} & 13.9 & 18.9 & 24.0 & 28.9\\
                
                Ubteacher~\cite{liu2021unbiased} & 16.0 & 20.0 & 27.1 & 28.0\\
                
                Noisy Boundaries~\cite{wang2022noisy} & 17.1 & 22.1 & 29.0 & 32.4\\
                
                PAIS~\cite{Hu2023pseudolabel} & 18.0 & 22.9 & 29.2 & 32.8 \\
                
                GuidedDistillation~\cite{berrada2023guided} & 23.0~\small{\color{blue}({+10.9})} & 30.8~\small{\color{blue}({+12.0})} & 33.1~\small{\color{blue}({+5.7})} & 35.6~\small{\color{blue}({+6.0})}\\

                \midrule

                PL-DC~(Ours) & \textbf{27.6}~\small{\color{blue}({+15.5})} & \textbf{33.1}~\small{\color{blue}({+14.3})} & \textbf{35.8}~\small{\color{blue}({+8.4})} & \textbf{37.6}~\small{\color{blue}({+8.0})}\\

                \bottomrule
            \end{tabular}
        \end{adjustbox}
    \end{center}
    \vspace{-5pt}
    \caption{
        Comparison with other SSIS on Cityscapes.
    }
    \label{table_sota_cityscapes}
    \vspace{-15pt}
\end{table}

\begin{figure*}[!t]
\centering
\includegraphics[width=1.0\linewidth]{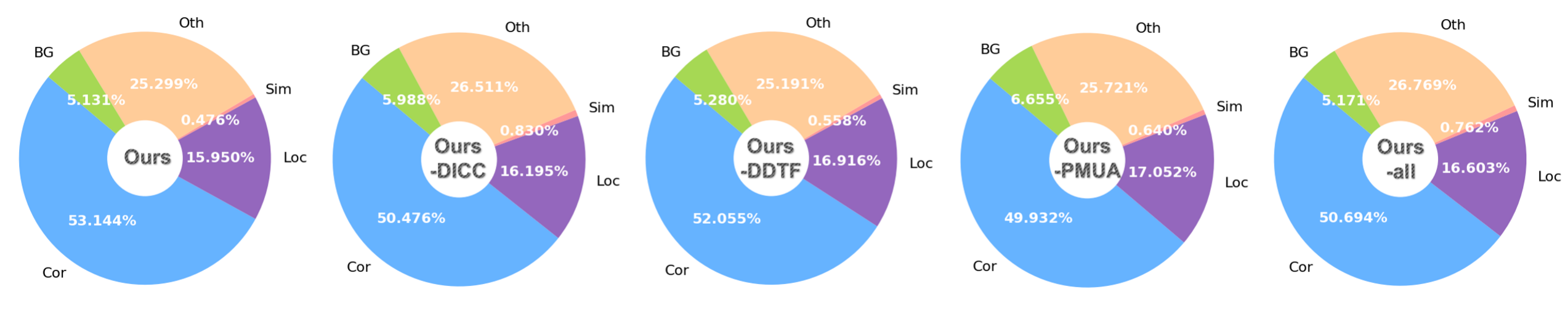}
\vspace{-20pt}
\caption{\textbf{Segmentation Analysis.} We randomly sampled $1k$ images from the COCO \textit{train2017} dataset to analysis the segmentation results. They are categorized into $5$ types: correct segmentation~(Cor), poor localization~(Loc), confusion with similar objects~(Sim), confusion with objects of other categories~(Oth), and confusion with the background~(BG).}
\label{fig:segmentation_analysis}
\vspace{-5pt}
\end{figure*}
\subsection{Abalation Study}
We conduct ablation studies on the proposed modules and hyper-parameters using the COCO dataset with $1\%$ labeled data over $73K$ iterations~($10$ epochs).
\textbf{Modules Validity} We ablate the Decoupled Dual-Threshold Filtering~(DDTF), Dynamic Instance Category Correction~(DICC), and Pixel-level Mask Uncertainty-Aware~(PMUA) modules, as depicted in Tab.\,~\ref{table_abalation}. 
Removing DDTF and replacing it with a coupled score threshold~($0.9\times0.85=0.765$) diminishes the $m$AP, $AP_m$, and $AP_l$, yet enhances $AP_s$.
This phenomenon occurs because, for medium and large objects, the competition between mask quality and class quality prevents the coupled score threshold filtering mechanism from simultaneously evaluating both the class quality and mask quality of an instance effectively.
Conversely, for small objects where the area is limited, class quality predominates in determining pseudo-label quality.
DDTF’s fixed mask quality threshold, which adversely affects small object quality, warrants further investigation.
Removing DICC results in a notable reduction in the $AP_s$ and $AP_m$, likely due to their smaller visual features and higher susceptibility to classification errors.
The removal of PMUA leads to a significant drop in the $AP_m$, attributable to the fact that the uncertainty area in medium objects represents a larger fraction of their total area.
The combined removal of all modules results in a less marked decline in overall $m$AP than removing PMUA alone, suggesting a balanced compromise between object classification and mask segmentation capabilities.
\begin{table}[t]
    \begin{center}
        \begin{adjustbox}{max width=0.48\textwidth}
            \begin{tabular}{l|llll}
                \toprule
                
                 & $m$AP & $AP_s$ & $AP_m$ & $AP_l$\\
                
                \midrule

                PL-DC~(Ours) & \textbf{21.6} & 7.0 & \textbf{21.4} & \textbf{35.2} \\

                \midrule
                
                - DDTF & 21.1~\tiny{\color{red} ($\downarrow$ 0.5) } & \textbf{7.3}~\tiny{\color{blue} ($\uparrow$ 0.3) } & 20.4~\tiny{\color{red} ($\downarrow$ 1.0) } & 34.9~\tiny{\color{red} ($\downarrow$ 0.3) } \\

                - DICC & 20.8~\tiny{\color{red} ($\downarrow$ 0.8) } & 6.2~\tiny{\color{red} ($\downarrow$ 0.8) } & 20.5~\tiny{\color{red} ($\downarrow$ 0.9) } & 35.0~\tiny{\color{red} ($\downarrow$ 0.2) } \\

                - PMUA & 20.4~\tiny{\color{red} ($\downarrow$ 1.2) } & 6.3~\tiny{\color{red} ($\downarrow$ 0.7) } & 20.0~\tiny{\color{red} ($\downarrow$ 1.4) } & 34.6~\tiny{\color{red} ($\downarrow$ 0.6) } \\

                \midrule

                - all above & 20.7~\tiny{\color{red} ($\downarrow$ 0.9) } & 6.4~\tiny{\color{red} ($\downarrow$ 0.6) } & 20.4~\tiny{\color{red} ($\downarrow$ 1.0) } & 34.7~\tiny{\color{red} ($\downarrow$ 0.5) } \\
                
                \bottomrule
            \end{tabular}
        \end{adjustbox}
    \end{center}
    \vspace{-5pt}
    \caption{
        \textbf{Ablation study~(model trained 70k) on COCO 1\%.} ``-" means remove module. - DDTF: remove DDTF and replace it with a coupled score threshold~(0.765) filtering. We evaluate the standard COCO metrics: $m$AP, $AP_s$ for small objects, $AP_m$ for medium objects, and $AP_l$ for large objects.
    }
    \label{table_abalation}
    \vspace{-10pt}
\end{table}
\begin{table}[t]
    \begin{center}
        \begin{adjustbox}{max width=1.0\linewidth}
            \begin{tabular}{ccc}
                \begin{subtable}[t]{0.4\linewidth}
                    \centering
                    \begin{tabular}{ll|c}
                        \toprule
                         $m_t$ & $c_t$ & $mAP$ \\
                        \midrule
                        - & - & 21.1 \\
                        0.7 & 0.7 & 20.3 \\
                        0.8 & 0.7 & 20.4 \\
                        0.9 & 0.7 & 21.1 \\
                        0.9 & 0.8 & 21.5 \\
                        0.9 & 0.85 & \textbf{21.6} \\
                        0.9 & 0.9 & 20.1 \\
                        \bottomrule
                    \end{tabular}
                    \caption{Different mask quality threshold $m_t$ and class quality threshold $c_t$ in DDTF.}                    
                \end{subtable} &
                \begin{subtable}[t]{0.3\linewidth}
                    \centering
                    \begin{tabular}{l|c}
                        \toprule
                         $\alpha$ & $mAP$ \\
                        \midrule
                        0.5 & 19.9 \\
                        0.7 & 20.0 \\                        
                        0.9 & 20.0 \\
                        0.99 & 20.2 \\
                        0.999 & 21.0 \\
                        0.9996 & \textbf{21.6} \\
                        0.9999 & 19.1 \\
                        \bottomrule
                    \end{tabular}
                    \caption{Different EMA rate $\alpha$.}
                \end{subtable} &
                \begin{subtable}[t]{0.3\linewidth}
                    \centering
                    \begin{tabular}{l|c}
                        \toprule
                         $\lambda$ & $mAP$ \\
                        \midrule
                        0.5 & 21.1 \\
                        1 & \textbf{21.6} \\
                        2 & 18.6 \\
                        4 & 10.0 \\
                        8 & 6.6 \\
                        \bottomrule
                    \end{tabular}
                    \caption{Different weight $\lambda$ for the unsupervised loss $\mathcal{L}_\mathrm{unsup}$.}
                \end{subtable}
            \end{tabular}
        \end{adjustbox}
    \end{center}
    \vspace{-5pt}
    \caption{Hyper-parameters in Our PL-DC.}
    \label{table_abalation_hyper-parameters}
    \vspace{-15pt}
\end{table}
\begin{table}[h]
    \centering
    \begin{adjustbox}{max width=1.0\linewidth}
        \begin{tabular}{l|lccc}
            \toprule
            & & $1\%$ & $5\%$ & $10\%$ \\
            
            \midrule
            
            &PAIS~\cite{Hu2023pseudolabel} & 21.1 & 29.3 & 31.0 \\
            &GuidedDistillation~\cite{berrada2023guided} & 21.5 & 29.9 & 35.0 \\
            &PL-DC (Ours)&25.1&33.0&36.5\\
            
            \midrule
        
            a&GuidedDistillation + pred maskIoU~\cite{Hu2023pseudolabel}&21.7&30.5&35.6\\
        
            b&GuidedDistillation + DDTF&23.5&30.9&35.7\\
        
            \midrule
        
            c&GuidedDistillation + BMP~\cite{wang2022noisy}&22.4&30.7&35.5\\ 
        
            d&GuidedDistillation + PMUA&23.1&31.3&35.7\\ 
        
            \midrule

            e&GuidedDistillation + PLePI~\cite{haghighi2023pseudo}&23.8&31.2&35.7\\
            f&GuidedDistillation + DICC&24.5&31.6&35.9\\

            g&PAIS + PLePI~\cite{haghighi2023pseudo}&23.3&30.5&33.6\\
            h&PAIS + DICC&24.1&30.8&34.0\\

            \bottomrule
        \end{tabular}
    \end{adjustbox}
    \vspace{-5pt}
    \caption{Experiments on the compatibility of modules.}
    \label{table_general_capabilities}
    \vspace{-15pt}
\end{table}

\textbf{Hyper-parameters Tuning} We abalate mask quality threshold $m_t$ and class quality threshold $c_t$ in DDTF, EMA rate $\alpha$ and the unsupervised loss $\mathcal{L}_\mathrm{unsup}$ weight $\lambda$ in Tab.\,\ref{table_abalation_hyper-parameters}.
From Tab.\,\ref{table_abalation_hyper-parameters}~(a), we observed three key phenomena. 
1) In DDTF, when $c_t$ is set low, the model's performance becomes more sensitive to $m_t$. As depicted in Fig.\,\ref{fig:pseudo_label_threshold} (b)(c), this sensitivity arises because the mask quality modeling does not accurately reflect the IoU relationship with the oracle GT. Conversely, class quality more accurately mirrors this relationship;
2) A low combination of $m_t$ and $c_t$ introduces more noisy pseudo-labels, resulting in reduced pseudo-label accuracy and a corresponding decline in the model’s $m$AP;
3) Conversely, setting both $m_t$ and $c_t$ too high leads to over-filtering of pseudo-labels, resulting in diminished pseudo-label recall and a decrease in $m$AP.
From Tab.\,\ref{table_abalation_hyper-parameters}~(b), it is evident that a smaller EMA rate $\alpha$ results in lower $m$AP, suggesting that the student model significantly influences the teacher model with each iteration, potentially propagating the negative effects of noisy pseudo-labels. Optimal performance is achieved at an EMA rate of $0.9996$. However, further increasing $\alpha$ slows down updates to the teacher model, as it relies predominantly on its previous weights.
As shown in Tab.\,\ref{table_abalation_hyper-parameters}~(c), the model performs optimally when the unsupervised loss weight $\lambda$ is set to $1.0$. Increasing this weight further leads to a sharp decline in $m$AP, indicating a detrimental effect on model performance.
\subsection{Compatibility with Other SSIS Methods}
We investigated the compatibility of our proposed modules with existing SSIS frameworks. 
To compare our DDTF with the mask IoU prediction used in PAIS~\cite{Hu2023pseudolabel} for evaluating mask quality, we added a mask IoU prediction branch to GuidedDistillation~\cite{berrada2023guided} in experiment 'a' of Tab.\,\ref{table_general_capabilities}, thus implementing PAIS's dual-threshold filtering strategy. 
Compared to experiment 'b', our DDTF shows greater effectiveness under sparse labeled data, indicating it is less prone to overfitting due to limited annotations.
Experiments 'c' and 'd' compare the BMP from Noisy Boundaries~\cite{wang2022noisy} with our proposed PMUA.
Our PMUA proves to be more general than BMP, which uses the distance to object boundaries as a weighting map.
Furthermore, since our DICC is the first to introduce CLIP~\cite{radford2021learning} for pseudo-label correction in SSIS, we compared it with PLePI~\cite{haghighi2023pseudo}, an SSOD method that uses CLIP by modeling the joint probability distribution of the teacher's and CLIP's predicted class distributions.
This comparison was conducted on both PAIS based on Faster R-CNN and GuidedDistillation based on Mask2Former.
As shown in experiments 'e', 'f', 'g', and 'h' of Tab.\,\ref{table_general_capabilities}, our DICC outperforms PLePI in both models.
We believe this is because CLIP is aligned at the image level and is significantly affected by the lack of context at the instance level.
Our DICC effectively balances correcting the model's class predictions and mitigating CLIP's noise by dynamically controlling the probability weights of CLIP predictions.
\subsection{Qualitative Analysis}
In Fig.\,\ref{fig:segmentation_analysis}, we analyze the impact of each module of our PL-DC on instance segmentation results.
We randomly sampled $1k$ images from the COCO \textit{train2017} dataset and categorized the instance segmentation results into five types: \textit{correct segmentation}~(Cor), where the mask IoU exceeds $0.5$ with any ground truth~(GT) and the category matches; \textit{poor localization}~(Loc), where the mask IoU ranges between $0$ and $0.5$ with any GT and the category matches; \textit{confusion with similar objects}~(Sim), where the mask IoU is above $0.5$ with any GT and the category is similar~(belonging to the same superclass in COCO); \textit{confusion with objects of other categories}~(Oth), where the mask IoU is above $0.5$ with any GT but the category differs; and \textit{confusion with the background}~(BG), where the mask IoU is $0$ with any GT.
From this analysis, we can draw four conclusions: (1) Removing DICC increases the proportion of Oth and Sim errors, suggesting that DICC somewhat mitigates confusion between objects. (2) Removing PMUA leads to a higher occurrence of Loc and BG errors, indicating that PMUA enhances the mask quality for objects. (3) Removing DDTF impacts BG, Sim, and Loc, as DDTF regulates the quality of pseudo labels at the instance level. (4) Most errors originate from confusion with objects of different categories. We propose that integrating more sophisticated category correction techniques to address inaccurate classifications could further enhance the performance of our PL-DC. 
%
\section{Conclusion}
\label{sec:conclusion}
In this paper, we introduced a novel Pseudo-Label Quality Decoupling and Correction~(PL-DC) framework to address the critical challenges in semi-supervised instance segmentation~(SSIS). 
PL-DC effectively mitigates the issues of pseudo-label noise at the instance, category, and pixel levels through three innovative modules: Decoupled Dual-Threshold Filtering, Dynamic Instance Category Correction, and Pixel-level Mask Uncertainty-Aware loss.
Extensive experiments on COCO and Cityscapes demonstrated the significant performance improvements achieved by PL-DC, setting new state-of-the-art SSIS results. 
%
{
    \small
    \bibliographystyle{ieeenat_fullname}
    \bibliography{main}
}

\clearpage
\setcounter{page}{1}
\setcounter{section}{0}
\setcounter{figure}{0}
\setcounter{table}{0}
\setcounter{equation}{0}
\renewcommand{\thefigure}{\Roman{figure}}
\renewcommand{\thetable}{\Roman{table}}
\renewcommand{\theequation}{\roman{equation}}
\maketitlesupplementary
\appendix

\section{Overview}
In this supplementary material, we provide additional experimental results and analyses.
\begin{itemize}
    \item More Implementation Details.
    \item Pixel-Level Mask Uncertainty-Aware Validity Proof.
    \item The Impact of CLIP Recognition Capability on DICC
    \item More Quantitative and Qualitative Analyses.
\end{itemize}
\section{More Implementation Details}
\label{sec_sup:implementation_details}
\subsection{Model and Training}
Our implementation builds upon Mask2Former~\cite{cheng2021mask2former} with ResNet-50 pretrained on ImageNet with fully supervision for fair comparison, which is coded in Detectron2 framework~\cite{wu2019detectron2}. 
Consistent with Ubteacher~\cite{liu2021unbiased}, PAIS~\cite{Hu2023pseudolabel} and GuidedDistillation~\cite{berrada2023guided}, we used a so-called ``burn-in" stage to train our models on only labeled data. After that, run a teacher-student ``mutual learning" on labeled and unlabeled data.
By default, all experiments are conducted on a single machine equipped with four 3090 GPUs, each with 24 GB of memory. For optimization, we utilize AdamW~\cite{loshchilov2017decoupled} with a learning rate and weight decay both set at 0.0001. Due to limited GPU memory, all network backbones are frozen.
The Clip~\cite{radford2021learning}, trained in a contrastive learning manner on a dataset of about $400$ million image-text pairs collected on the Internet, used in Dynamic Instance Category Correction~(DICC) uses R50 as the backbone. It receives a $224\times224$ resolution image and text with a maximum length of $77$ tokens as input.  
%
\subsection{Hyper-parameters and Data augmentation}
For the setting of hyper-parameters, as shown in Tab.\,\ref{tab:sup_hyper_parameters}, we use mask quality threshold $m_t = 0.9$ and class quality threshold $c_t = 0.85$ to filter the pseudo-labels. We use $\alpha = 0.9996$ for EMA and $\lambda = 1$ for the unsupervised loss Lunsup. 
On the COCO setup, we pre-train the teacher model in ``burn-in" stage about $20k$ iterations. Afterward, the student model is initialized with the parameters of the teacher model, and run teacher-student ``mutual learning". The total training iterations for each semi-supervised learning are all $360K$~($50$ epochs), with batch sizes consistently comprising $8$ labeled and $8$ unlabeled images unless otherwise specified. 
On Cityscapes setup, the hyper-parameters mirror those of the COCO configuration, except the total training duration is reduced to $180k$ iterations, and the batch sizes are halved to $8$.
On Ablation Study setup, the hyper-parameters mirror those of the COCO configuration, except the total training duration is reduced to $73k$ iterations, and the batch sizes are halved to $8$.
For the Data augmentation, as shown in Tab.\,\ref{tab:sup_data_aug}, we apply random horizontal flip, scale jittering and fixed size crop as weak augmentations for the teacher model, while the student model receives strong augmentations including horizontal flip, scale jittering, fixed size crop, color jittering, grayscale, gaussian blur, and CutOut~\cite{devries2017improved}.
\begin{table*}[ht]
    \begin{center}
        \begin{adjustbox}{max width=1.0\linewidth}
            \begin{tabular}{l|l|c|c|c}
                \toprule
                Hyper-parameter & Description & COCO & Cityscapes & Ablation \\
                $m_t$ & Mask quality threshold & 0.9 & 0.9 & 0.9 \\
                $c_t$ & Class quality threshold & 0.85 & 0.85 & 0.85 \\
                $\alpha$ & EMA rate & 0.9996 & 0.9996 & 0.9996 \\
                $\lambda$ & Unsupervised loss weight & 1.0 & 1.0 & 1.0 \\
                $b_l$ & Batch size for labeled data & 8 & 4 & 4 \\
                $b_u$ & Batch size for unlabeled data & 8 & 4 & 4 \\
                burn-in & Train model only on label data & 20000 & 20000 & 10000 \\
                max iteration & Maximum number of iterations for model training & 368750 & 180000 & 73750 \\
                $\gamma$ & Learning rate & 0.0001 & 0.0001 & 0.0001 \\     
                \bottomrule
            \end{tabular}
        \end{adjustbox}
    \end{center}
    \caption{Hyper-parameters in our PL-DC.}
    \label{tab:sup_hyper_parameters}
\end{table*}
\begin{table*}[!htbp]
\begin{center}
\begin{adjustbox}{max width=1.0\textwidth}
\begin{tabular}{l|cccccccc}
\toprule

  &Avg P&Avg R&bear R&skis R&elephant R&knife R&bottle R&mouse R\\

\midrule

CLIP& 74.8 & 50.5 & 98.0 & 96.7 & 96.6 & 1.1 & 2.4 & 2.3 \\

\bottomrule
\end{tabular}
\end{adjustbox}
\end{center}
\caption{
CLIP's precision and recall for familiar and unfamiliar categories on COCO 2017 val.
}
\label{tab:sup_PR}
\end{table*}
\begin{table*}[ht]
    \begin{center}
        \begin{adjustbox}{max width=1.0\linewidth}
            \begin{tabular}{lcp{4cm}p{8cm}}
            \toprule
            Process & Probability & Parameters & Descriptions \\
            \midrule
            \multicolumn{4}{c}{Weak Augmentation} \\
            \midrule
            Horizontal Flip & 0.5 & - & None \\
            \midrule
            \multirow{4}{*}{Scale Jittering} & \multirow{4}{*}{1.0} & (min\_scale, max\_scale, target\_height, target\_width) = (0.1, 2.0, 1024, 1024) & Takes target size as input and randomly scales the given target size between ``min\_scale" and ``max\_scale". It then scales the input image such that it fits inside the scaled target box, keeping the aspect ratio constant. \\
            \midrule
            \multirow{4}{*}{FixedSizeCrop} & \multirow{4}{*}{1.0} & (height, width) = (1024, 1024) & If ``crop\_size" is smaller than the input image size, then it uses a random crop of the crop size. If ``crop\_size" is larger than the input image size, then it pads the right and the bottom of the image to the crop size. \\
            \midrule
            \multicolumn{4}{c}{Strong Augmentation} \\
            \midrule
            Horizontal Flip & 0.5 & - & None \\
            \midrule
            \multirow{4}{*}{Scale Jittering} & \multirow{4}{*}{1.0} & (min\_scale, max\_scale, target\_height, target\_width) = (0.1, 2.0, 1024, 1024) & Takes target size as input and randomly scales the given target size between ``min\_scale" and ``max\_scale". It then scales the input image such that it fits inside the scaled target box, keeping the aspect ratio constant. \\
            \midrule
            \multirow{4}{*}{FixedSizeCrop} & \multirow{4}{*}{1.0} & (height, width) = (1024, 1024) & If ``crop\_size" is smaller than the input image size, then it uses a random crop of the crop size. If ``crop\_size" is larger than the input image size, then it pads the right and the bottom of the image to the crop size. \\
            \midrule
            \multirow{4}{*}{Color Jittering} & \multirow{4}{*}{0.8} & (brightness, contrast, saturation, hue) = (0.4, 0.4, 0.4, 0.1) & Brightness factor is chosen uniformly from [0.6, 1.4], contrast factor is chosen uniformly from [0.6, 1.4], saturation factor is chosen uniformly from [0.6, 1.4], and hue value is chosen uniformly from [-0.1, 0.1]. \\
            \midrule
            Grayscale & 0.2 & None & None \\
            \midrule
            \multirow{2}{*}{GaussianBlur} & \multirow{2}{*}{0.5} & (sigma\_x, sigma\_y) = (0.1, 2.0) & Gaussian filter with $\sigma_x = 0.1$ and $\sigma_y = 2.0$ is applied. \\
            \midrule
            \multirow{2}{*}{CutoutPattern1} & \multirow{2}{*}{0.7} & scale=(0.05, 0.2), ratio=(0.3, 3.3) & Randomly selects a rectangle region in an image and erases its pixels. \\
            \midrule
            \multirow{2}{*}{CutoutPattern2} & \multirow{2}{*}{0.5} & scale=(0.02, 0.2), ratio=(0.1, 0.6) & Randomly selects a rectangle region in an image and erases its pixels. \\
            \midrule
            \multirow{2}{*}{CutoutPattern3} & \multirow{2}{*}{0.3} & scale=(0.02,0.2),ratio=(0.05, 0.8) & Randomly selects a rectangle region in an image and erases its pixels. \\
            \bottomrule
            \end{tabular}
        \end{adjustbox}
    \end{center}
    \caption{Details of data augmentation in our PL-DC.}
    \label{tab:sup_data_aug}
\end{table*}

\section{Pixel-Level Mask Uncertainty-Aware\\ Validity Proof}
\label{sec_sup:PMUA_Proof}
In the main paper, we define the pixel-level mask binary cross-entropy loss for training the student model on unlabeled data:
\begin{equation}
    \begin{split}
        \mathcal{L}_{mask}^{unsup} =& -\frac{1}{QHW} \sum_{k=1}^{Q} \sum_{i=1}^{HW} (1-u_{\hat{\sigma}(k)}^i) [ M_{\hat{\sigma}(k)}^i \log (t_k^i) \\ +
        & (1 - M_{\hat{\sigma}(k)}^i) \log (1 - t_k^i) ]
    \end{split}
\end{equation}
$u_{\hat{\sigma}(k)}^i$ represents the uncertainty of the pseudo-label of the pixel mask at position $i$. A larger value indicates increased noise in the pseudo-label $M_{\hat{\sigma}(k)}^i$.
For simplicity, we assume the student network output as $t=\theta x$, where $\theta$ is the learnable parameter and $x$ is the input image.
We need to compute the derivative of $\mathcal{L}_{mask}^{unsup}$ with respect to $\theta$ to better understand the effect of noise on the gradient descent algorithm updating $\theta$.
According to the chain rule of derivation:
\begin{equation}
    \frac{\partial \mathcal{L}_{mask}^{unsup}}{\partial \theta} = \frac{\partial \mathcal{L}_{mask}^{unsup}}{\partial t_k^i} \cdot \frac{\partial t_k^i}{\partial \theta},
\end{equation}
\begin{equation}
    \begin{split}
    \frac{\partial \mathcal{L}_{mask}^{unsup}}{\partial t_k^i} &= -\frac{1}{QHW} \sum_{k=1}^{Q} \sum_{i=1}^{HW} \left( 1 - u_{\hat{\sigma}(k)}^i \right) \\
    &\cdot \left[ \frac{M_{\hat{\sigma}(k)}^i}{t_k^i} - \frac{1 - M_{\hat{\sigma}(k)}^i}{1 - t_k^i} \right],
    \end{split}
\end{equation}
\begin{equation}
    \frac{\partial t_k^i}{\partial \theta} = x_k^i,
\end{equation}
\begin{equation}
    \begin{split}
    \frac{\partial \mathcal{L}_{mask}^{unsup}}{\partial \theta} &= -\frac{1}{QHW} \sum_{k=1}^{Q} \sum_{i=1}^{HW} \left( 1 - u_{\hat{\sigma}(k)}^i \right) \\
    &\cdot \left[ \frac{M_{\hat{\sigma}(k)}^i}{t_k^i} - \frac{1 - M_{\hat{\sigma}(k)}^i}{1 - t_k^i} \right] \cdot x_k^i.
    \end{split}
\end{equation}
From the form of the derivative, we observe that when $u_{\hat{\sigma}(k)}^i$ is larger, indicating increased noise for $M_{\hat{\sigma}(k)}^i$, the factor $(1-u_{\hat{\sigma}(k)}^i)$ approaches zero. Consequently, the overall value of the derivative $\frac{\partial \mathcal{L}_{mask}^{unsup}}{\partial \theta}$ decreases. This demonstrates that higher noise levels reduce the influence on the derivative of the loss function, thereby minimizing the impact on the $\theta$ update during the gradient descent process.
It can be concluded that the larger the noise, the smaller the derivative, the smaller the influence on the loss function, and thus the smaller the influence on the update of the parameter  $\theta$ in the gradient descent algorithm. This phenomenon can be understood as the disturbance effect on the parameter $\theta$ is reduced when the noise is large, and the update of the gradient descent is more stable.
\section{The Impact of CLIP Recognition Capability on DICC}
\label{sec_sup:impact_clip}
\begin{figure*}[!t]
\centering
\includegraphics[width=1.0\linewidth]{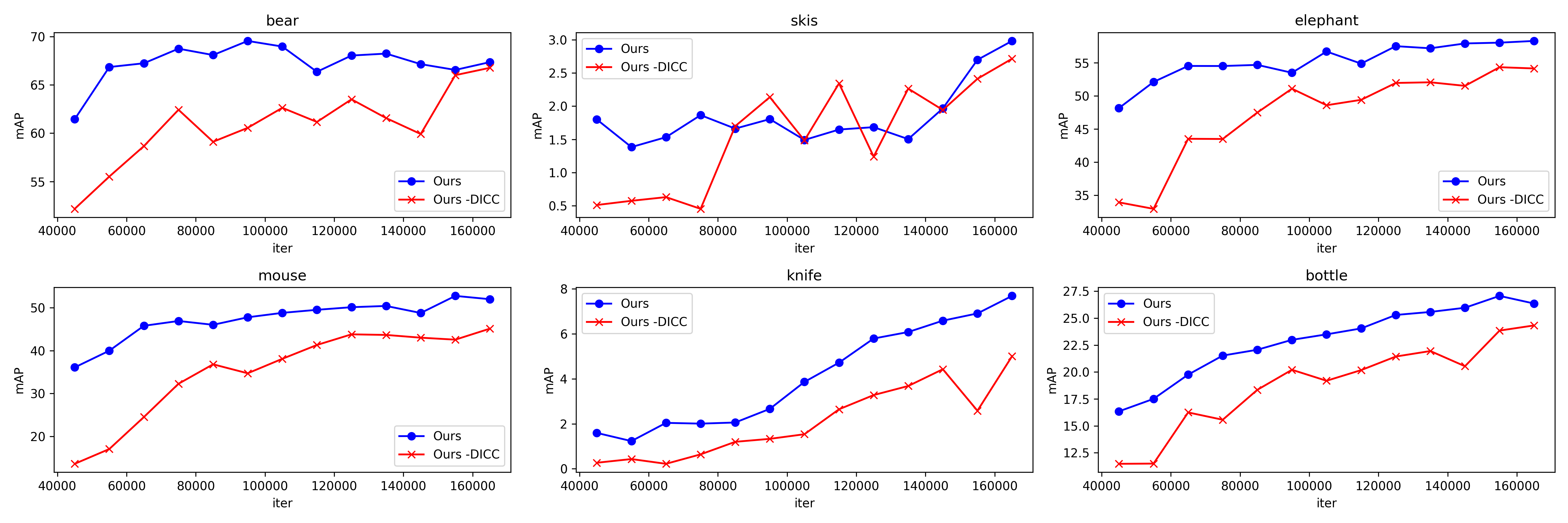}
\caption{DICC can handle classes that CLIP is not familiar with. \textbf{Better View in Zoom.}}
\label{fig:sup_clip_impact}
\end{figure*}

To analyze the impact of CLIP's recognition ability on our proposed Dynamic Instance Category Correction (DICC) module, in Tab.\,\ref{tab:sup_PR}, we utilize ground truth (GT) masks of objects from the COCO 2017 val dataset to extract corresponding visual patches, which are then fed into CLIP for category prediction. 
We calculate recall and precision by comparing the predicted categories with the GT classes. 
The average precision and recall across the 80 categories are 74.8 and 50.5, respectively.
The three categories with the highest recall (the ones most familiar to CLIP) are bear, skis, and elephant, with recall rates of 98.0, 96.7, and 96.6, respectively. 
The three categories with the lowest recall (the ones least familiar to CLIP) are knife, bottle, and mouse, with recall rates of 1.1, 2.4, and 2.3, respectively.
In Fig.\,\ref{fig:sup_clip_impact}, we visualize the convergence curves for training these categories under COCO 10\%. We observe several interesting points: 1) Our DICC not only accelerates convergence but also improves the final mAP by 1–6. 2) For categories that are less familiar to CLIP, CLIP overcomes initial biases at the beginning of training. Eq.\,\ref{eq:dicc_1}-\ref{eq:dicc_3} demonstrate that the weight for CLIP's class prediction gradually decays from 0.5 to 0, which results in DICC relying more on the teacher model rather than CLIP, as the teacher’s reliability increases during training. 3) For CLIP-familiar categories such as skis, while CLIP can improve mAP, the final mAP is still not high. This is because skis are often covered by snow, and our model only segments the portions not covered by snow, while the GT mask uses coarse annotations that label both the skis and the snow, leading to inaccurate evaluation.
These observations confirm that the effectiveness of DICC is not solely dependent on CLIP; it is also influenced by our dynamic weighting algorithm and the capabilities of the teacher model. 
DICC is only utilized during training, adding no extra parameters during inference, so the required resources remain the same. In fact, DICC can incorporate any VLM for improved category recognition, albeit with a higher forward time cost during training. For instance, we tested LLaVA-1.6, which achieved an average precision of 56.7 and a recall of 90.1. However, due to the excessive inference time, it cannot be directly used for training our model. We plan to explore distillation techniques from powerful VLMs to lightweight VLMs in the future.
\section{More Quantitative and Qualitative Analyses}
\begin{figure*}[ht!]
    \centering
    \begin{subfigure}[b]{0.33\textwidth}
        \centering
        \includegraphics[width=\textwidth]{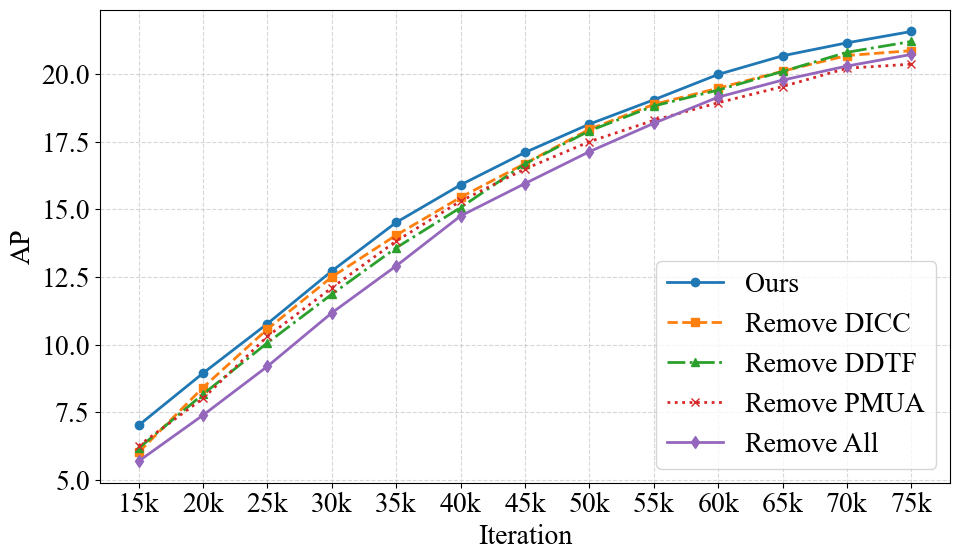}
        \caption{AP Convergence Curve.}
    \end{subfigure}
    \hfill
    \begin{subfigure}[b]{0.33\textwidth}
        \centering
        \includegraphics[width=\textwidth]{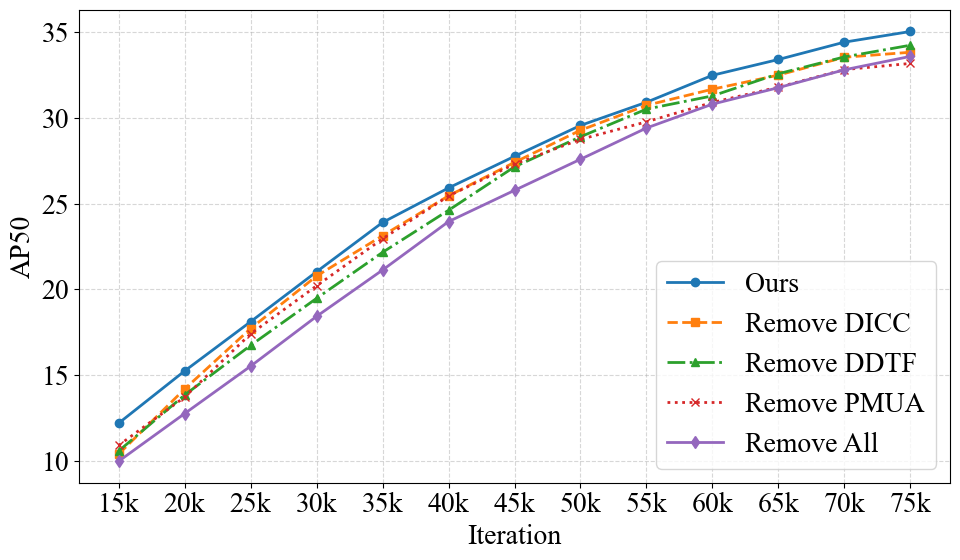}
        \caption{AP50 Convergence Curve.}
    \end{subfigure}
    \hfill
    \begin{subfigure}[b]{0.33\textwidth}
        \centering
        \includegraphics[width=\textwidth]{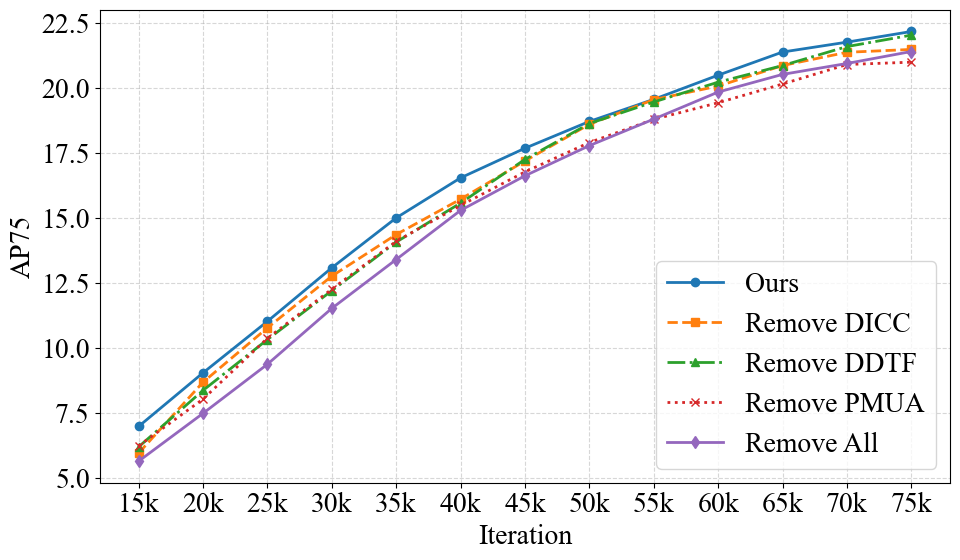}
        \caption{AP75 Convergence Curve.}
    \end{subfigure}
    \begin{subfigure}[b]{0.33\textwidth}
        \centering
        \includegraphics[width=\textwidth]{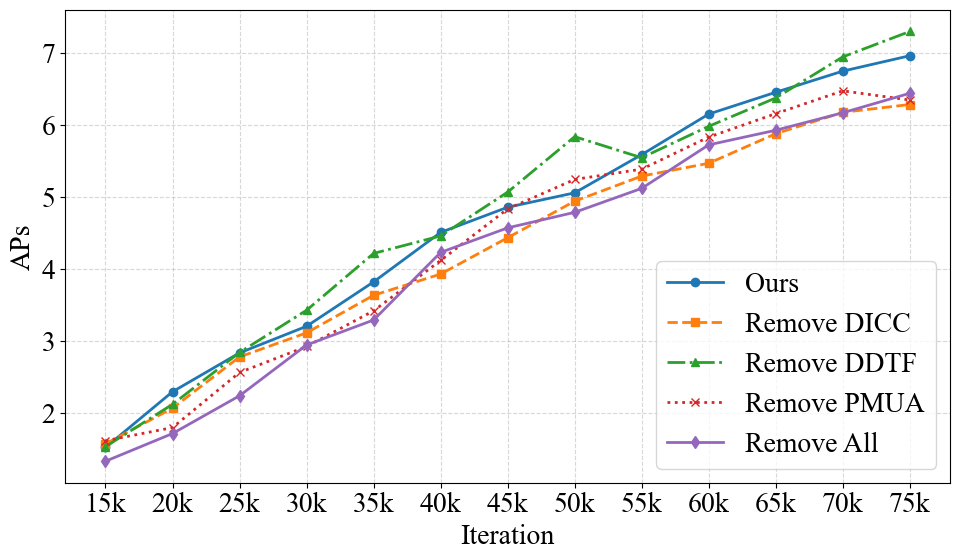}
        \caption{$\text{AP}_s$ Convergence Curve.}
    \end{subfigure}
    \hfill
    \begin{subfigure}[b]{0.33\textwidth}
        \centering
        \includegraphics[width=\textwidth]{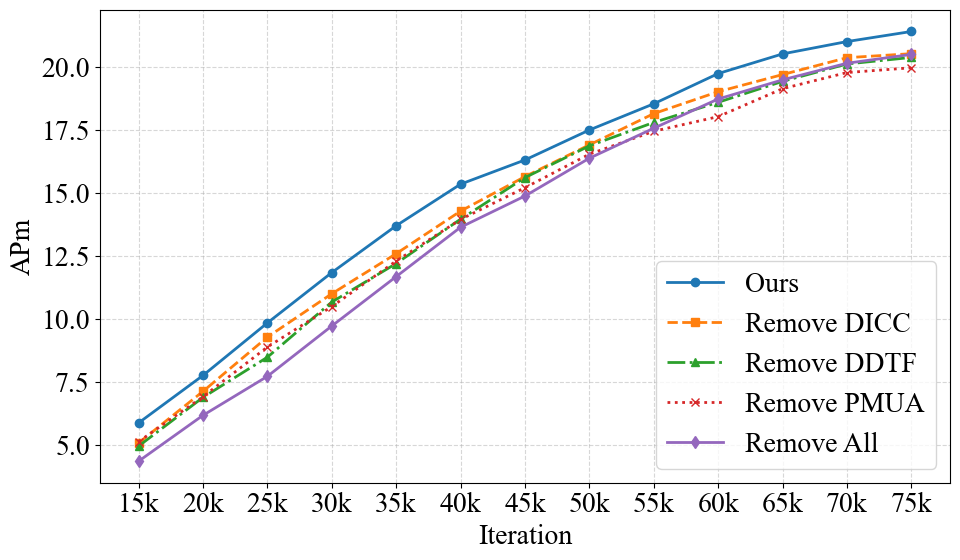}
        \caption{$\text{AP}_m$ Convergence Curve.}
    \end{subfigure}
    \hfill
    \begin{subfigure}[b]{0.33\textwidth}
        \centering
        \includegraphics[width=\textwidth]{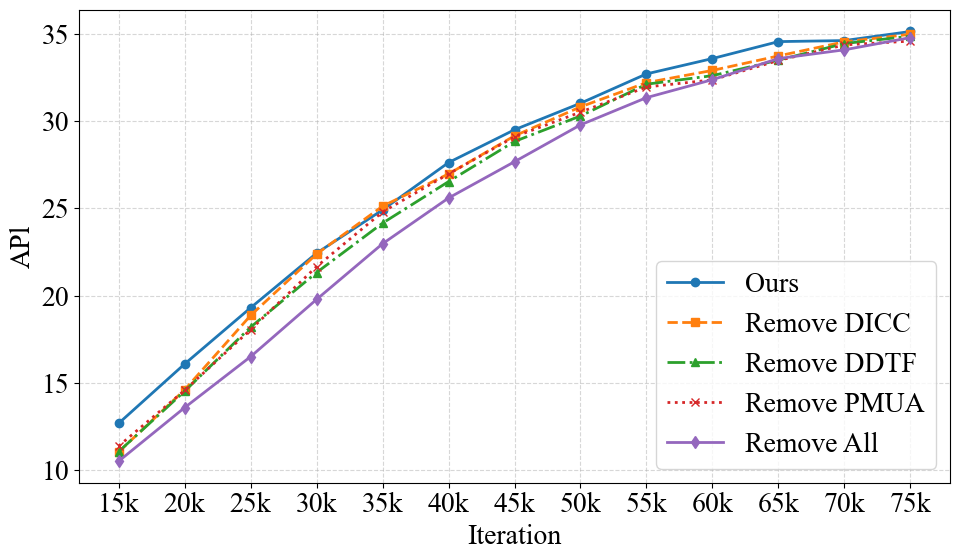}
        \caption{$\text{AP}_l$ Convergence Curve.}
    \end{subfigure}
    \caption{The impact of removing each module on the performance convergence of PL-DC. We train all models on COCO \textit{train2017} dataset with 1\% labeled data and the rest as unlabeled data over 73K iterations~(10 epoches), and test all models on COCO \textit{val2017}. \textbf{Better View in Zoom.}}
    \label{fig:AP_splits}
\end{figure*}

\begin{figure*}[!t]
\centering
\includegraphics[width=1.0\linewidth]{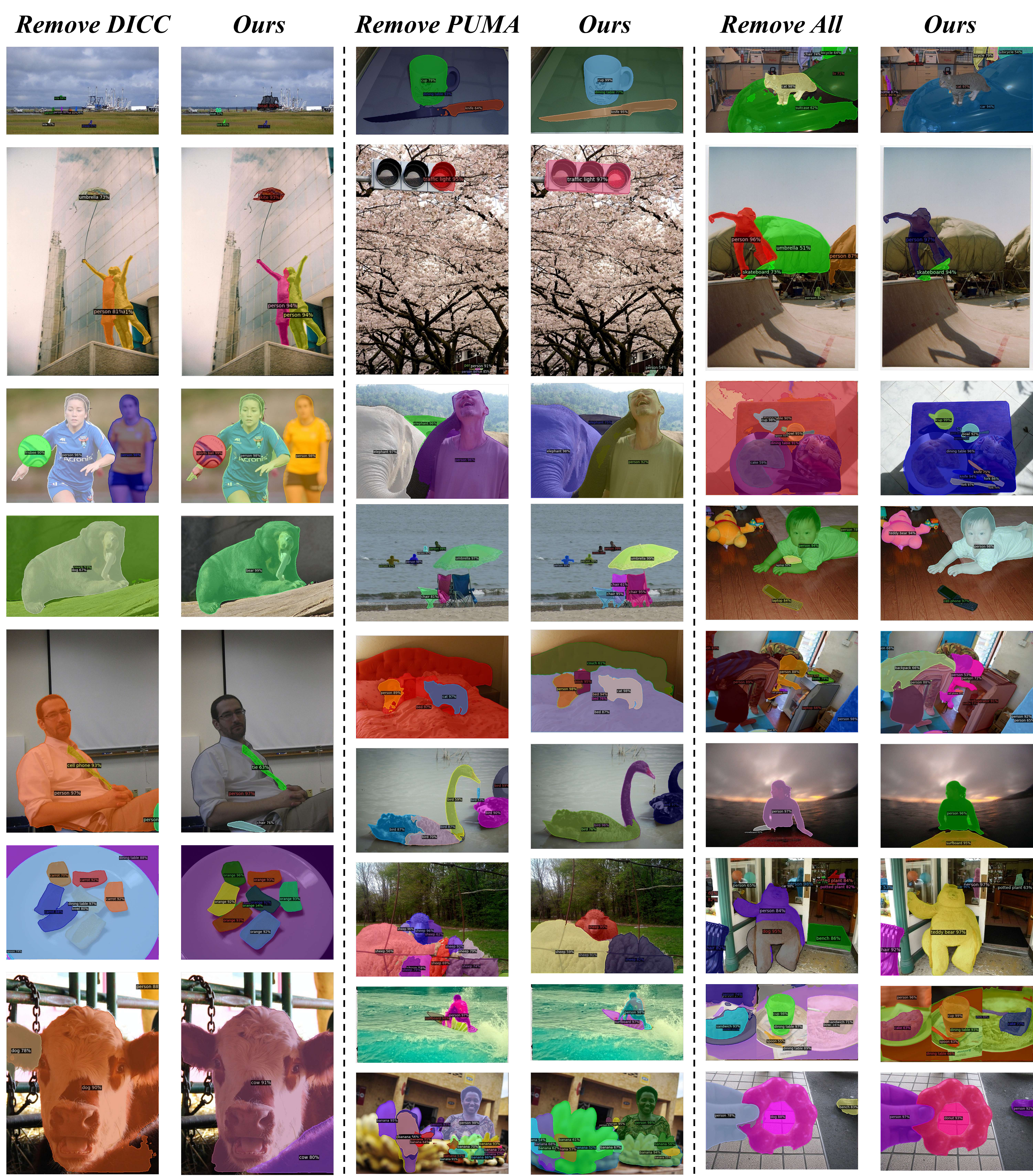}
\caption{Visualization of the impact of DICC and PMUA on PL-DC segmentation results. \textbf{Better View in Zoom.}}
\label{fig:sup_abalation}
\end{figure*}

In Fig.\,\ref{fig:AP_splits}, we visualized the impact of removing each modules of PL-DC on the convergence of AP, AP50, AP75, $\text{AP}_s$, $\text{AP}_m$, and $\text{AP}_l$.
We discovered that removing the Decoupled Dual-Threshold Filtering~(DDTF) and replacing it with a coupled score threshold~($0.9 \times 0.85 = 0.765$) is disadvantageous for the training of small objects in the early stages, but as training progresses into the middle and later stages, it becomes more beneficial for small object training. This is because, in the early stages of training, both the quality of mask and class determine the quality of small objects, making DDTF more effective than a fixed threshold filtering mechanism. However, in the mid to late stages of training, for small objects with limited area, the quality of the class plays a leading role in determining the quality of pseudo-labels, while the mask, due to their small area, make DDTF's fixed mask quality threshold inappropriate for evaluating small object mask, warrants further investigation.
Removing the Dynamic Instance Category Correction~(DICC) module primarily results in poor training outcomes for small objects, as their visual features are smaller and more prone to classification errors. The DICC module effectively addresses this issue of category confusion, making it crucial for accurately classifying small objects.
The removal of the Pixel-Level Mask Uncertainty-Aware~(PMUA) module leads to difficulties in training objects of medium size, as uncertain areas within these medium objects constitute a larger proportion of their total area. This highlights the critical importance of the PMUA in training mask uncertainty, particularly for objects where the area of uncertainty is substantial relative to their overall size.
In Fig.\,\ref{fig:sup_abalation}, we visualize the impact of removing different modules of PL-DC on the final segmentation results.
Removing the Dynamic Instance Category Correction~(DICC) module primarily leads to errors in the categorization of objects in instance segmentation, such as classifying a "bird" as a "kite," a "kite" as an "umbrella," a "bear" as a "dog," and a "cow" as a "dog." Removing the Pixel-Level Mask Uncertainty-Aware~(PMUA) module mainly results in poor masks for segmented objects, such as only part of the handle being segmented without the blade for a "knife," only one light being segmented for a "traffic light," and excessive segmentation including other sheep for a "sheep." Removing both DICC and PMUA leads to both categorization errors and poor masks in instance segmentation, such as the front and glass of a "car" being segmented into two parts, with the front being classified as "suitcase" and the glass as "tv," and a "teddy bear" doll being segmented into two parts, with the upper body classified as "person" and the lower body as "dog." These visualizations highlight that our DICC module is aimed at solving the problem of confusion in predicting instance categories, while the PMUA module is focused on addressing the uncertainty in predicting instance masks.

\end{document}